\documentclass{article}
\usepackage{arxiv}
\usepackage[utf8]{inputenc} 
\usepackage[T1]{fontenc}    
\usepackage{hyperref}       
\usepackage{url}            
\usepackage{booktabs}       
\usepackage{amsfonts}       
\usepackage{nicefrac}       
\usepackage{microtype}      
\usepackage{lipsum}
\usepackage{graphicx}
\usepackage{amssymb}
\graphicspath{ {./images/} }
\usepackage{cite}
\usepackage{amsmath}
\usepackage{multirow}

\title{RORPCap: Retrieval-based Objects and Relations Prompt for image captioning}

\author{
 Jinjing Gu \\
  School of Information Science and Engineering\\
  Yunnan University\\
   \\
  \texttt{jinjinggu@ynu.edu.cn} \\
   \And
    Tianbao Qin \\
  School of Information Science and Engineering\\
  Yunnan University\\
   \\
  \texttt{tantianbao@stu.ynu.edu.cn} \\
  \And
 Yuanyuan Pu \\
  School of Information Science and Engineering\\
  Yunnan University\\
     \\
  \texttt{yuanyuanpu@ynu.edu.cn} \\
    \And
 Zhengpeng Zhao \\
  School of Information Science and Engineering\\
  Yunnan University\\
     \\
  \texttt{zhpzhao@ynu.edu.cn} \\
}

\begin{document}
\maketitle
\begin{abstract}
Image captioning aims to generate natural language descriptions for input images in an open-form manner. To accurately generate descriptions related to the image, a critical step in image captioning is to identify objects and understand their relations within the image. Modern approaches typically capitalize on object detectors or combine detectors with Graph Convolutional Network~(GCN). However, these models suffer from redundant detection information, difficulty in GCN construction, and high training costs. To address these issues, a Retrieval-based Objects and Relations Prompt for Image Captioning~(RORPCap) is proposed, inspired by the fact that image-text retrieval can provide rich semantic information for input images. RORPCap employs a Objects and Relations Extraction Model to extract object and relation words from the image. These words are then incorporate into predefined prompt templates and encoded as prompt embeddings. Next, a Mamba-based mapping network is designed to quickly map image embeddings extracted by CLIP to visual-text embeddings. Finally, the resulting prompt embeddings and visual-text embeddings are concatenated to form textual-enriched feature embeddings, which are fed into a GPT-2 model for caption generation. Extensive experiments conducted on the widely used MS-COCO dataset show that the RORPCap requires only 2.6 hours under cross-entropy loss training, achieving 120.5\% CIDEr score and 22.0\% SPICE score on the ‘Karpathy’ test split. RORPCap achieves comparable performance metrics to detector- and GCN-based models with the shortest training time and demonstrates its potential as an alternative for image captioning.
\end{abstract}


\section{Introduction}

\begin{figure}[h]
\centering
\centerline{\includegraphics[width=10.5cm]{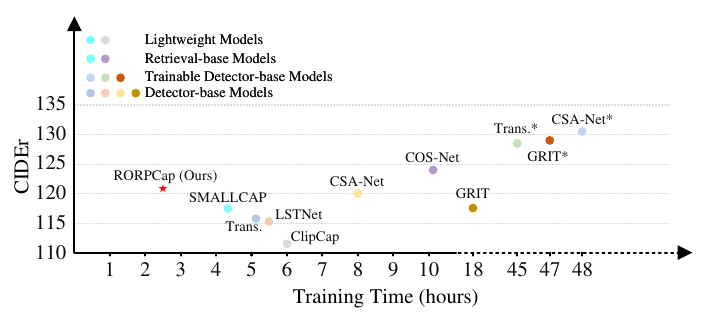}}
\caption{Comparison of training time and CIDEr scores for each model. *: Detector is trainable. All results were trained on the MS-COCO dataset using a single RTX 4090 and optimized with cross-entropy loss.}
\label{fig2}
\end{figure}

Image captioning is a high-level semantic understanding task in computer vision, aiming to generate a visually-grounded and linguistically coherent sentence, which covers most semantics in an image that are worthy of mention. But achieving this goal faces the issue of great disparities between the visual and language domains~\cite{yan2020deep}. To accurately generate descriptions related to the image, a critical step in image captioning is to identify objects and understand their relations within the image. Modern approaches typically capitalize on object detectors or combine detectors with Graph Convolutional Network~(GCN). However, as shown in Fig.~\ref{fig2}, these models suffer from redundant detection information, difficulty in GCN construction, and high training costs. 

\renewcommand{\figurename}{Fig.}

\begin{figure}[h]
\centering
\centerline{\includegraphics[width=9.5cm]{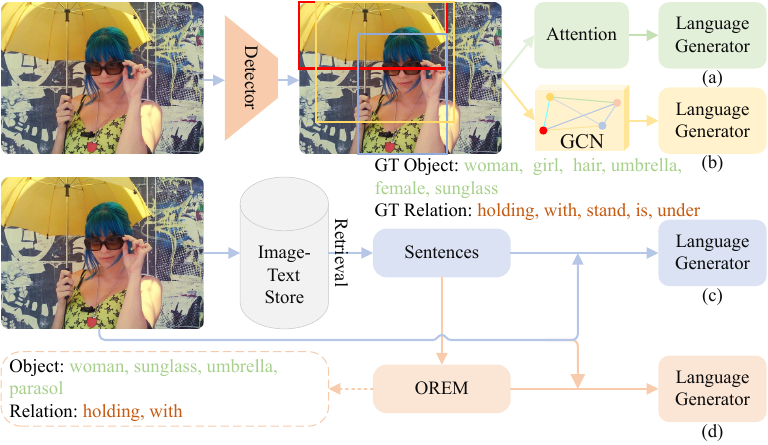}}
\caption{Image captioning based on (a) detector and attention, (b) detector and GCN, (c) cross-model retrieval, (d) our Objects and Relations Extraction Model (OREM).}
\label{fig1}
\end{figure}

As illustrated in Fig.~\ref{fig1}, to obtain features of objects, some detector-based methods\cite{MultimodalTransformer-11, 12, li2023csa-13, hossen2024attribute-28} use detector to encode objects, as shown in Fig.~\ref{fig1} (a). These methods perform well, but detectors may be constrained by the object categories in pre-trained models\cite{li2022comprehending-26}. For instance, the DETR \cite{zhu2020deformable-14}, trained on the MS-COCO dataset with only 80 categories, struggles to recognize object like "sunglass" and relations. Using more diverse detectors\cite{anderson2018bottom-50} may introduce redundant detections and increase computational cost. To capture relations in the image, some works use GCN \cite{kipf2016semi-15} to introduce semantic and spatial relations \cite{yao2018exploring-16, hong2021variational-17, xiao2024sentinel-56}, as shown in Fig.~\ref{fig1} (b). GCN-based methods can also achieve good results, but they suffer with the challenges of complex GCN construction and high computational costs \cite{ghandi2023deep-18}. 
\renewcommand{\figurename}{Fig.}

\cite{feng2019unsupervised} pointed out that image captioning systems require additional textual concepts\cite{4, 5} as supervised information to ensure the relevance of the generated descriptive sentence to the input image. Textual concepts can be object words (e.g., chair, computer, people) and relation words (e.g., standing, sitting, with), which carry important visual information in the language domain. Previous work\cite{6} on sentence grammar also demonstrated the importance of nouns, verbs, gerunds, and other words in the structure, meaning, and fluency of sentences. These suggest that textual modality information is helpful to generate a semantically and grammatically correct descriptive sentence for the input image. To capture textual concepts, some retrieval-based image captioning models\cite{ramos2023smallcap-8, ramos2023retrieval-9} mainly use the input image to search for captions from a repository. As shown in Fig.~\ref{fig1} (c), retrieval methods are used to search for captions similar to the input image, which serve as prompt information to guide the generation of descriptions along with the input image. However, using multiple complete sentences may add redundancy, affecting description quality and increasing training time. To solve the above problem, this study proposed Retrieval-based Objects and Relations Prompt for Image Captioning~(RORPCap). As shown in Fig.~\ref{fig1} (d), a Objects and relations Extraction Model~(OREM) is designed to obtain object and relation words within the image, which are mainly represented by nouns, verbs, gerunds, and prepositions, which can effectively avoid the troubles caused by detectors and GCNs.

Merely using object and relation words within the image is insufficient for accurately generating captions; visual information must also be taken into account. With the advent of vision-language model CLIP \cite{radford2021learning-19}, which is capable of mapping vision and text into a shared vector space. \cite{mokady2021clipcap-62, ramos2023smallcap-8, kim2025vipcap-58} leverage the abundant visual-textual representation of CLIP to achieve model lightweighting. To reduce model training time, we also utilize CLIP. However, the representations obtained from CLIP remain independent of the latent space of the language model. We need a bridge to connect the vision and text domains. Inspired by ClipCap \cite{mokady2021clipcap-62}, this work employed a mapping network built on Transformer \cite{vaswani2017attention-27} as a bridge. Our RORPCap chooses to construct the mapping network based on Mamba \cite{gu2023mamba-41} as the bridge. Mamba is a new selective state space model where the state transition matrix can be dynamically adjusted based on the current input. This allows the model to more flexibly capture key information in the sequences, and the network provides faster computation speeds for longer sequences.

As shown in Fig.~\ref{fig3}, the RORPCap leverages object and relation words about the image obtained from OREM and fills these words into a template to convert into prompt embedding using the tokenizer of the language model GPT-2 \cite{radford2019language-20}. To bridge the gap between the visual and textual modalities, a mapping network built on Mamba is used to map the image embeddings extracted by CLIP into visual-textual embeddings. Next, the prompt embeddings and visual-textual embeddings are concatenated to form a prefix. During training, the prefix is connected to the ground-truth~(GT) embeddings and then fed into the GPT-2. During inference, the language model generates descriptions word by word, conditioned on the prefix. Extensive experiments are conducted on the MS-COCO \cite{lin2014microsoft-21} dataset and the nocaps \cite{agrawal2019nocaps-22} dataset, which aims to measure generalization to unseen classes and concepts. The results show that our RORPCap model has the potential to serve as an alternative to detector- and GCN-based image captioning models. RORPCap aims to explore an alternative approach to detector-based and GCN-based image captioning models. RORPCap does not require a detector or construct a GCN, which helps reduce training parameters and the model's training time. The contributions of this study can be summarized as follows:
\begin{itemize}
\item The OREM is proposed for extracting image objects and relation words, and constructing a prompt template to enhance image captioning performance. This is a reusable integrated module for other works.
\item A fast and simple mapping network is designed to bridge the gap between the vision and textual domains based on Mamba. We conducted ablation experiments to verify the performance differences between the Transformer and the Mamba as mapping networks from vision and language.
\item Extensive experiments are conducted on the standard image captioning benchmarks of the MS-COCO and achieve performance comparable to state-of-the-art models on the ‘Karpathy’ split. Moreover, RORPCap is the model with the shortest training time for comparable performance metrics and demonstrates well zero-shot capability on the nocaps dataset.
\end{itemize}

The paper is structured as follows: Section \ref{sec:2} presents prior studies. Section \ref{sec:3} details presents our framework and method. Section \ref{sec:4} showcases experiments and result analysis. Qualitative results in Section \ref{sec:5}, while Section \ref{sec:6} concludes with a summary and suggestions for future research.

\section{Relted Works} \label{sec:2}
To extract objects and relations from the image, existing image captioning models can be roughly classified into three types: detector-based, GCN-based, and retrieval-based image captioning. This section briefly reviews some existing works that are closely related to this study.

\subsection{\textbf{Detector-based Image Captioning}}
To obtain the features of objects in the image, some detector-based methods\cite{10, MultimodalTransformer-11, 12, li2023csa-13, hossen2024attribute-28, parvin2023image-29, cai2024top-30, al2024nposc-31} use pre-trained detectors. \cite{hossen2024attribute-28} takes a detector to extract prominent objects features from prominent regions, then inputs them into an attribute-driven filtering captioning network to predict the most pertinent attributes in accordance with the textual context. \cite{parvin2023image-29} employs a detector to extract objects features and proposes an improved double-attention framework to enhance the performance of the image captioning. Although these image captioning models that freeze the detector are able to effectively extract object features, they still face limitations of vocabulary diversity, speed, and performance. With the development of Vision Transformers, some end-to-end image captioning models based on detectors\cite{li2023csa-13, nguyen2022grit-33} have been proposed. CSA-Net \cite{li2023csa-13} uses DETR\cite{zhu2020deformable-34} to extract visual region features and builds object and relation semantic cues from the pre-trained RegionCLIP text encoder\cite{zhong2022regionclip-35}. GRIT\cite{nguyen2022grit-33} utilizes a deformable DETR as an object detector and considers the complementary visual region and grid features as input to encode visual relations. These detector-based image captioning models achieve good performance by applying the powerful detection capabilities, but detectors struggle to capture object relations in the image, which introduces interference in the caption generation and increases the training time of the model. 

\subsection{\textbf{GCN-base Image Captioning}}
To capture the relations in the image, many researchers \cite{yao2018exploring-16, yang2019auto-36, zhang2021consensus-37, tripathi2021sg2caps-38, liu2021region-39, tong2024reversegan-40} use detectors combined with GCN\cite{hong2021variational-17} to extract high-level semantics and visual relations in the image. \cite{yang2019auto-36} constructs a Scene Graph Autoencoder~(SGAE), which applies directed scene graph representations to capture the complex relations in the image. \cite{tripathi2021sg2caps-38} utilizes spatial positions of objects and interaction labels of man-made objects to capture the relations in the image. \cite{liu2021region-39} constructs object and region-level graphs and optimizes them using the Spatial GCN Interaction module to capture relations both within and across regions and objects. To inclusion complex semantic relations in images, \cite{tong2024reversegan-40} introduces an adversarial network combined with GCN for image captioning. GCN-based image captioning models can effectively identify relations between objects in an image and yield excellent results, but the complexity of graph construction remains a key challenge\cite{ghandi2023deep-18}. Instead, the proposed RORPCap method applies the Objects and Relations Extraction Model~(OREM) to obtain the objects and relations in the image as an alternative to using detectors and GCNs. This approach allows us to avoid using detectors and GCNs, reduce computational burden, and speed up model training.

\subsection{\textbf{Retrieval-based Image Captioning}}
To achieve a lightweight model, some retrieval-based image captioning models\cite{ramos2023smallcap-8, ramos2023retrieval-9, kuznetsova2014treetalk-25, li2022comprehending-26, kim2025vipcap-58} are proposed to capture textual concepts. For instance, \cite{ramos2023smallcap-8} fills sentences similar to the input image into a designed prompt template and inputs them into a language model to generate the final description. \cite{ramos2023retrieval-9} uses a pre-trained visual-language encoder to encode the input image and retrieves sentences to obtain multimodal representations. \cite{li2022comprehending-26} retrieves sentences related to the image from an external corpus, filters out the irrelevant semantic words, and infers the missing relevant words visually grounded in the image. Furthermore, different with other retrieval-based models, \cite{kim2025vipcap-58} leverages the retrieved text with image information as visual prompts to enhance the ability of the model to capture relevant visual information. These methods directly use the retrieved sentences, which can contain redundant information such as adjectives, adverbs, and positional words, interfering with the captioning generation and negatively impacting its quality. However, the large amount of information to be processed leads to longer training times. Our RORPCap model extracts object and relation words from the retrieved sentences that match the input image, which helps to reduce training time.

\subsection{\textbf{Mapping networks}}

To enhance training efficiency, recent works\cite{mokady2021clipcap-62, manas2022mapl-64, alayrac2022flamingo-65} suggest focusing on learning a mapping network. ClipCap\cite{mokady2021clipcap-62} uses CLIP encoding as a prefix to the caption by employing a simple mapping network, and then fine-tunes a language model to generate the image captions. But the results have been underwhelming. Additionally, ClipCap only uses image features without additional textual concepts as prompts, leading to insufficient expressiveness of the model. \cite{manas2022mapl-64} leverages a pre-trained vision encoder and a pre-trained language model, and learns a mapping network to convert visual features into token embeddings. \cite{alayrac2022flamingo-65} shares a core Transformer \cite{vaswani2017attention-27} stack and a fixed number of learned constant embeddings. These works all use the Transformer as the mapping network. In the proposed RORPCap model, the Mamba \cite{gu2023mamba-41} is chose as our mapping network. Mamba is a novel selective state space model that dynamically adjusts the state transition matrix based on the current input. This enables the model to more effectively capture key information within sequences, while the network also offers faster computation for longer sequences.

\section{Method} \label{sec:3}

\begin{figure*}[h]
\centerline{\includegraphics[width=1\linewidth]{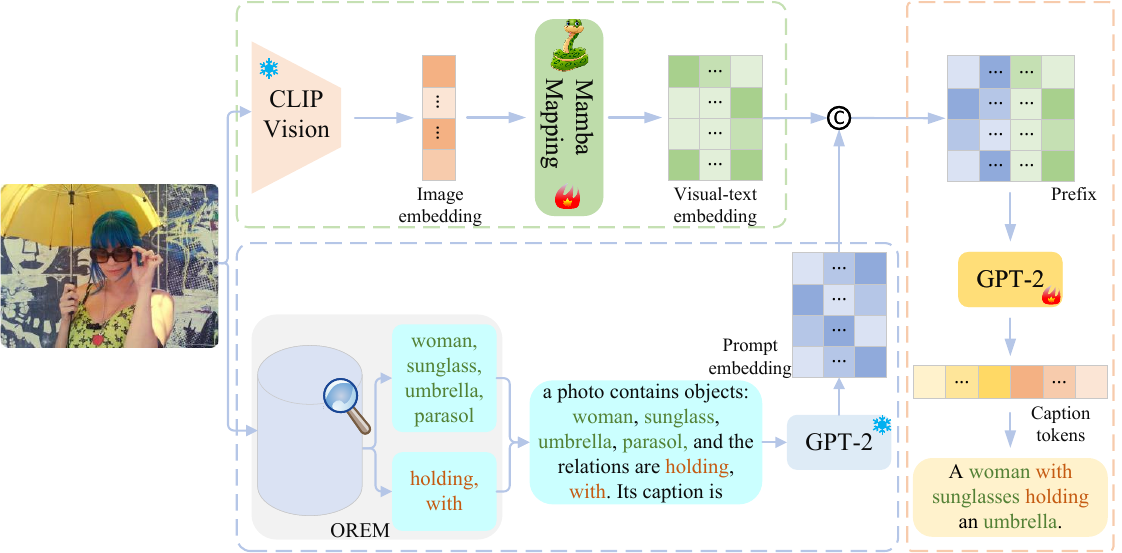}}
\caption{An overview of the proposed RORPCap. 1) We first use the OREM (see Fig.~\ref{fig4}) to retrieve the sentences most similar to the input image and extract object and relation words. Then, we fill these words into the constructed prompt template and utilize GPT-2 tokenizer to project the textual concepts to the prompt embedding. 2) We use the pretrained CLIP visual encoder to obtain image embedding and feed the image embedding into the Mamba mapping network to obtain the visual-text embedding. 3) We concatenate the prompt embedding and visual-text embedding to obtain the prefix embedding, and then feed the prefix embedding into the language model GPT-2.} 
\label{fig3}
\end{figure*}

Fig.~\ref{fig3} provides an overall illustration of the RORPCap method, which mainly consists of OREM and a mapping network. To obtain textual concepts, we use the OREM in Fig.~\ref{fig4} to retrieve the sentences most similar to the input images and extract object and relation words. We use GPT-2 (small) as our language model and utilize its tokenizer to project the prompt and caption to sequence of embedding $\text{\textit{p}}^{j}$, $\text{\textit{c}}^{j}$. To extract visual information from the input image $\text{\textit{x}}^{j}$, we use the pretrained CLIP model's visual encoder to obtain image embedding. Next, the image embedding is put into a mapping network \textit{M} to get a visual-text embedding. We can get a prefix by connecting the prompt embedding with the visual-text embedding, which is expressed as follows: 
\begin{equation}
    v_{1}^{j}, \ldots, v_{n}^{j}=\operatorname{\text{\textit{GPT}}} \left(p^{j}\right)\textit{+}\textit{M}\left[\operatorname{\text{\textit{CLIP}}}\left(x^{j}\right)\right]
\end{equation}

After Eq.(1), a \textit{n} length prefix vector can be obtained, where the embedding vector $v_{n}^{j}$ has the same dimension as the caption embedding $\text{\textit{c}}^{j}$. During training, the prefix embedding vector $v_{n}^{j}$ should concatenate with the caption embedding $\text{\textit{c}}^{j}$, as expressed by the following formula:
\begin{equation}
    \text { \textit{Conne} }^{j}=v_{1}^{j}, \ldots, v_{n}^{j}, c_{1}^{j}, \ldots, c_{l}^{j}
\end{equation}

In the model training phase, the $\text{\textit{Conne}}^{\textit{j}}$ is fed into the language model, where the input information is used as a prior condition in an autoregressive manner to predict subsequent tokens that are consistent with the image. Then, we train the mapping component \textit{M} and fine-tune the language model using the simple, yet effective, cross-entropy loss:
\begin{equation}
    \mathcal{L}_{X}=-\sum_{j=1}^{N} \sum_{i=1}^{l} \log p_{\theta}\left(c_{i}^{j} \mid v_{1}^{j}, \ldots, v_{n}^{j}, c_{1}^{j}, \ldots, c_{i-1}^{j}\right)
\end{equation}
where $\theta$ is the learnable parameters in the model, \textit{N} denotes the total number of datasets, \textit{l} is the maximal length of caption.

\subsection{\textbf{Objects and Relations Extraction Model}}
\begin{figure*}[h]
\centerline{\includegraphics[width=1\linewidth]{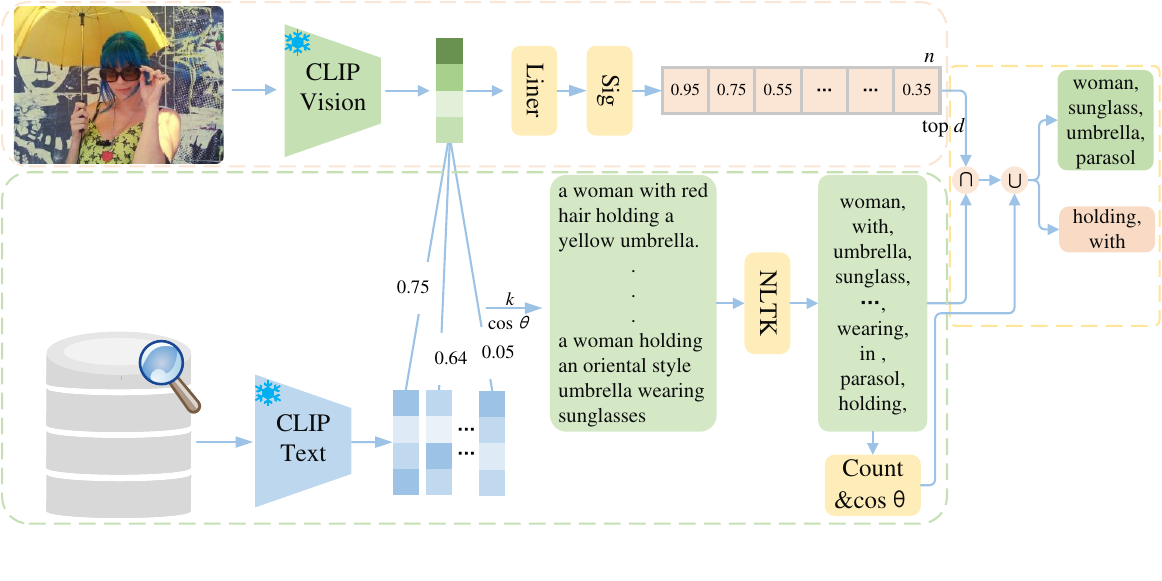}}
\caption{An overview of the proposed OREM. 1) We use a linear layer and an activation layer to map the image features to \textit{n} high-frequency words and obtain the corresponding scores. 2) We use NLTK to decompose the sentence into nouns, gerunds, verbs, and prepositions. 3) We take the intersection of the results from 1) and 2).} 
\label{fig4}
\end{figure*}

Directly applying the entire sentence as a prompt\cite{ramos2023smallcap-8, ramos2023retrieval-9} may introduce redundant information. To address this, as shown in Fig.~\ref{fig4}, we process the retrieved sentences and extract important information from them in the OREM. 

As shown in Fig.~\ref{fig4}, the CLIP visual and text encoders are used to map the visual and text into a shared vector space. We encode an input image $I$ and the datastore contents $S$, then use nearest neighbor search based on cosine similarity to retrieve the \textit{k} most similar sentences from the datastore (excluding the ground-truth of the input image), which is formulated as follows:
\begin{equation}
    \operatorname{\textit{Sim}}(I, S)=\frac{\mathbf{v}_{i} \cdot s_{k}}{\left\|\mathbf{v}_{i}\right\|\left\|s_{k}\right\|}
\end{equation}
where $\mathbf{v}_{i}$ is the feature of the input image, $s_{k}$ is the feature of the sentences in the datastore, the \textit{k} searched captions are \( S_k = \{ sent_1, sent_2, \dots, sent_k \} \). 

As mentioned in the analysis above, objects and relations in the image play an important role in generating descriptions. Our key idea is to extract these two types of words from \textit{k} sentences. Objects in the image are primarily represented as nouns, while relation words mainly appear as verbs, gerunds, prepositions, etc. We use a linear layer and an activation layer to map the image features to \textit{n} high-frequency words and obtain the corresponding scores. Then, we extract the top \textit{d} with scores greater than \textit{s} to form the set ${W_t}$. We use NLTK \cite{loper2002nltk-42} to tag part of speech in the retrieved sentences. To avoid interference from redundant information, such as adjectives and adverbs, we only extract object and relation words from the image, forming a word set ${W_s}$ as semantic clues. Next, we take the intersection of sets ${W_t}$ and ${W_s}$ to obtain the word set \( W_n = \{ w_1, w_2, \dots, w_n \} \). However, these words may not provide enough textual concepts, so we also incorporate word frequency and similarity statistics as supplements. Specifically, we perform a frequency analysis of object and relation words after part-of-speech tagging, selecting words whose object frequency is greater than the threshold \textit{o} and whose relation frequency is greater than the threshold \textit{r}. In particular, for objects, we also calculate the similarity between the objects with the input image to improve the accuracy of word selection. The sets of final object words are denoted as \( W_o = \{ o_1, o_2, \dots, o_n \} \) and the relation words are denoted as \( W_r = \{ r_1, r_2, \dots, r_j \} \). When the object and relation words are obtained, they are filled in the slots in a fixed prompt template of the following form: \textit{a photo contains objects:} \textit{${object_1}$}, \dots, \textit{${object_n}$}, \textit{and the relations are} \textit{${relation_1}$}, \dots, \textit{${relation_j}$}. \textit{Its caption is} \{mapping network output and GT\}. Our goal is to select \textit{n} objects and \textit{m} relation words that are in the image; for cases where there are fewer than \textit{n} or \textit{m}, we use \textit{null} symbol to fill the slots. We can refer to the prompts and captions as a sequence of tokens \( p^j = \{ p_1^j, p_2^j, \dots, p_n^j \} \) and \( c^j = \{ c_1^j, c_2^j, \dots, c_n^j \} \) where we pad the tokens to a maximal \textit{n}. The training objective of the model is as follows:
\begin{equation}
    \max _{\theta} \sum_{j=1}^{N} \log P_{\theta}\left[c_{1}^{j}, \ldots, c_{n}^{j} \mid\left(p^{j}+i^{j}\right)\right]
\end{equation}
where ${i}^{j}$ is the $i$-th image visual-text embedding obtained by the mapping network.

\subsection{\textbf{Mapping Network}}
Merely using object and relation words between objects in an image is insufficient to convey its primary information. It is also necessary to account for the contextual relations between these details, such as color, shape, size, and position, ensuring the description is logical and readable. Extracting information from the input image becomes particularly important. But vision and language domains operate in different vector spaces, our key idea is to explore a simple and fast method to connect the visual and text domains. A mapping network is used to process the image embeddings obtained from the CLIP visual encoder. The output of the mapping network is filled into the prompt template. We choose the Mamba model as the bridge connecting the visual domain and the textual domain. Its processing flow for sequences can be represented by the following formula:
\begin{equation}
    \Delta_{t}=\operatorname{softplus}\left(W_{\Delta} x_{t}+b_{\Delta}\right), B_{t}=W_{B} x_{t}+b_{B}, C_{t}=W_{C} x_{t}+b_{C}
\end{equation}
where \( \Delta _t \in \mathbb{R}^E \) is the discretization factor that controls the state update at time step $t$, and \( B_t, C_t \in \mathbb{R}^E \) are the projection parameters for dynamically adjusting the input and output sequence. \( W_{\Delta}, W_B, W_C \) are learnable weight matrices, and \( E \) represents the state dimension. \( \Delta_t \), \( B_t \), and \( C_t \) dynamically change with the input, enabling the model to selectively focus on important information. Eq.(6) generates input-dependent dynamic parameters through linear projection, enhancing the ability to model context. 

The continuous-time state-space model is integrated into the deep learning model, requiring discretization to adapt to the input sequence. Therefore, \( A \) and \( B \) are discretized using the zero-order hold method and the time scale parameter \( \Delta \), which is calculated as follows:
\begin{equation}
    \bar{A}_{t}=\exp \left(A \odot \Delta_{t}\right), \bar{B}_{t}=\left(\frac{\exp \left(A \odot \Delta_{t}\right)-I}{A}\right) \odot B_{t}
\end{equation}
where \( A \in \mathbb{R}^E \), $\odot$ denotes element-wise multiplication.

The recursive update of the hidden state \( h_t \) through a scanning mechanism can be represented as follows. 
\begin{equation}
    h_{t}=\bar{A}_{t} \odot h_{t-1}+\bar{B}_{t} \odot\left(W_{x} x_{t}\right)
\end{equation}
where the parameters $\bar{A}_{t}$ and $\bar{B}_{t}$ are used to control the fusion weights between the historical memory and the current input, while \( W_x \) is used to map the input \( x_t \) to the state dimension \( E \).

The model output can be calculated as follows:
\begin{equation}
    y_{t}=\left\langle C_{t}, h_{t}\right\rangle+D x_{t}=\sum_{i=1}^{N} C_{t, i} h_{t, i}+D x_{t}
\end{equation}
where \( C_t \) is the parameter influenced by the input, varying depending on the input, enhancing the output's expressive ability. \( D \) is the learnable residual weight matrix. Through the mapping network, we can quickly bridge the gap between the vision and language domains, which makes the training time of RORPCap very short.

\subsection{\textbf{Language Generator}}

We use GPT-2~(small) as the language model, which has been proven to work well in the image captioning field and is capable of generating rich and diverse text. The mapping network addresses the issue of the independence between the CLIP representations of images and the latent space of the language model. The mapping network allows the embeddings obtained from the CLIP visual encoder to be used in the language model, generating meaningful captions. 

Since the semantic information for generating the description is encapsulated in the prefix, we want to use the prefix as a prior condition to generate the text description. We can utilize an autoregressive language model to predict the next token. Thus, the objective can be described as follows:
\begin{equation}
    \max _{\theta} \sum_{k=1}^{N} \sum_{j=1}^{l} \log P_{\theta}\left(c_{j}^{k} \mid\left(p^{k}+i^{k}\right), c_{1}^{k}, \ldots, c_{j-1}^{k}\right)
\end{equation}

Each captioning dataset contains different styles, which may not align naturally with the pre-trained language model. Therefore, we fine-tune the decoder model while training the mapping network. We found that directly fine-tuning the decoder model would significantly increase the number of trainable parameters. Accordingly, we use a fine-tuning approach where certain layers are frozen. This not only results in more expressive outcomes and provides greater flexibility for the model but also saves computational resources and shortens training time.

\subsection{\textbf{Inference}}
During the inference, we first extract the pre-processed vocabulary data and fill it into the prompt templates, and then encode it using the language model's tokenizer. Next, we extract the visual-text embedding of the input image \( x \) using the CLIP encoder and the mapping network \( M \). Then, the text embedding is concatenated with the visual-text embedding to form the prefix, and the prefix is input into the decoder model. The decoder model generates a caption conditioned on the prefix, predicting subsequent tokens one by one based on the guidance of the language model output. For each token, the decoder outputs a probability distribution over all vocabulary tokens, which are used to determine the next one by employing the beam search.

\section{Experiments and Result Analysis } \label{sec:4}
In this section, we introduce the experiments, including datasets, evaluation metrics, implementation details, and results analysis.

\subsection{\textbf{Datasets}}
The MS-COCO dataset\cite{lin2014microsoft-21} is widely used in the image captioning field. This dataset contains 123,287 images, each annotated with at least 5 different descriptive sentences. To make comparisons with most image captioning models, we follow the partitioning standard set by Karpathy\cite{karpathy2015deep-43}, which divides the data into 5,000 images for validation, 5,000 images for testing, and 113,287 images for training. 

Since the MS-COCO dataset is limited to 80 categories, we further evaluate the generalization ability of RORPCap on the nocaps dataset\cite{agrawal2019nocaps-22}. The nocaps dataset is divided into three subsets: the in-domain, which contains categories similar to the MS-COCO dataset; the near-domain, which includes MS-COCO categories as well as some novel categories; and the out-of-domain, which contains completely novel categories compared to the MS-COCO dataset.

\subsection{\textbf{Evaluation Metrics}}
To evaluate the quality of the descriptions generated by the proposed RORPCap, we employ automatic evaluation metrics that are standard in the image captioning domain, including: BLEU(B)\cite{papineni2002bleu-44}, METEOR\cite{vedantam2015meteor-45}, ROUGH(R)\cite{lin2004rouge-46}, CIDEr(C)\cite{vedantam2015cider-47}, and SPICE(S)\cite{anderson2016spice-48}. BLEU is based on $n$-gram precision and measures the overlap between the generated text and reference text. METEOR considers precision, recall, and ${F_1}$-score, incorporating synonym and morphological matching. ROUGH is based on $n$-grams, word sequences, and lexical overlap. CIDEr combines TF-IDF weighted $n$-grams and emphasizes the importance of information, while SPICE is a semantic-based evaluation metric.

\subsection{\textbf{Implementation Details}}
In RORPCap, the visual encoder is CLIP-ViT-B/32 while the text decoder is GPT-2~(small), with both components leveraging pretrained weights from HuggingFace\cite{wolf2020transformers-49}. The visual encoder parameters are frozen during training, extracting 512-dimension visual features. The mapping network adopts the Mamba architecture with 10 stacked Mamba Blocks and a state space dimension of 16. The CLIP embeddings are transformed into 10 sequence lengths, with a dimension of 768. For the decoder, we fine-tune the pre-trained GPT-2~(small) to better adapt to the dataset's style. To save computational resources and shorten training time, we freeze layers 1 to 8. The model has a total of 12 layers.
For the retrieval part, we use the retrieval model weights provided in \cite{li2022comprehending-26}, and the total high-frequency words is 906. For each input image, we retrieve the top 7 semantically similar sentences. We extract the top 20 candidates with similarity scores exceeding 0.8 to form the set ${W_t}$. In word frequency statistics, the word frequency of object words is greater than four times, and the similarity with the input image is greater than 0.24. The frequency of relational words is the highest and is greater than twice. The objects are limited to a maximum of 6. The maximum number of relation words is set to 3. Our model uses a batch size of 40, a learning rate of 9e-6, and the optimizer is Adma\cite{kingma2014adam-66}. During the training phase, we perform 5,000 warm-up iterations. The model is trained with cross-entropy loss using PyTorch framework, running on a single NVIDIA GeForce RTX 4090 with a total of 6 training epochs.

\subsection{\textbf{Results Analysis}}
\subsubsection{\textbf{Evaluation Analysis}}
To validate the effectiveness of RORPCap in image captioning and assess its potential as an alternative to detector- and GCN-based models, we conducted comparative experiments with image captioning models based on detectors, GCNs, and retrieval methods, as shown in Table~\ref{T1}. 

\begin{table}
\caption{Performance comparison with detector-base, GCN-base, retrieval-base model on MS-COCO offline Karpathy test split. All of the detectors are frozen. Bold denotes the best results, while underline represents the second-best results, and (-) indicates unevaluated scores. *: Results computed by us. All values are reported as percentages~(\%).}
\renewcommand\arraystretch{1.25}
\centering
\setlength{\tabcolsep}{1.5mm}{
\resizebox{\linewidth}{!}{
\begin{tabular}{lc|cccccc}
\hline
Models                & Method    & B@1($\uparrow$)  & B@4($\uparrow$)  & M($\uparrow$) & R($\uparrow$) & C($\uparrow$) & S($\uparrow$) \\ \hline
Up-Down \cite{anderson2018bottom-50}      & Detector  & 77.2 & 36.2 & 27.0   & 56.4    & 113.5 & 20.3  \\
AoANet \cite{10}       & Detector  & 77.4 & 37.2 & 28.4   & 57.5    & 119.8 & 21.3  \\
MT \cite{MultimodalTransformer-11}           & Detector  & 76.2 & 36.6 & 28.3   & 56.8    & 117.1 & -     \\
X-Linear \cite{12}     & Detector  & 77.3 & 37.0 & 28.7   & 57.5    & 120.0 & 21.8  \\
Trans. \cite{cornia2020meshed-52}       & Detector  & 76.0 & 35.8 & 28.0   & 56.5    & 115.4 & 21.1  \\
GRIT \cite{nguyen2022grit-33}      & Detector  & 77.3 & 36.5 & 28.2   & 57.0    & 117.6 & 21.2  \\
LSTNet* \cite{ma2023towards-51}      & Detector  & 76.3 & 36.6 & 28.3   & 56.8    & 116.1 & 21.0  \\
CSA-Net \cite{li2023csa-13}      & Detector  & 78.1 & \underline{38.3} & 28.5   & \underline{58.0}    & 120.0 & 21.8  \\ 
SGO-F(ADF) \cite{hossen2024attribute-28}  & Detector  & 77.0 & 36.6 & 27.9   & 57.0    & 115.0 & 21.1  \\
NPoSC-A3 \cite{al2024nposc-31}     & Detector  & 78.0 & 37.0 & 27.5   & 56.8    & 115.7 & 20.8  \\ \hline
GCN-LSTM \cite{yao2018exploring-16}     & GCN       & 77.3 & 36.8 & 27.9   & 57.0    & 116.3 & 20.9  \\
SGAE \cite{yang2019auto-36}         & GCN       & 77.6 & 36.9 & 27.7   & 57.2    & 116.7 & 20.9  \\
PyAtt \cite{chen2022improving-53}        & GCN       & 78.3 & 37.9 & \textbf{28.9}   & \textbf{58.5}    & 119.6 & 21.9  \\
SGT \cite{huang2022improve-54}          & GCN       & 77.3 & 37.0 & 28.3   & 57.4    & 118.1 & -     \\
MCA \cite{zhao2022aligned-55}          & GCN       & \textbf{78.5} & 37.1 & 28.2   & 57.6    & 117.2 & 21.1  \\
SSVSGOAR+S \cite{xiao2024sentinel-56}   & GCN       & \underline{78.4} & \textbf{40.3} & \underline{28.8}   & \textbf{58.5}    & 118.8 & \textbf{22.0}  \\ \hline
UpDown+MA \cite{fei2021memory-57}    & Retrieval & 77.1 & 37.1 & 28.3   & 57.2    & 116.3 & 21.3  \\
AoANet+MA \cite{fei2021memory-57}    & Retrieval & 78.2 & 38.0 & 28.7   & 57.8    & \underline{121.0} & 21.8  \\
SMALLCAP* \cite{ramos2023smallcap-8}     & Retrieval & 77.4 & 36.2 & 27.6   & 56.7    & 117.7 & 21.0  \\
EXTRA \cite{ramos2023retrieval-9}         & Retrieval & -    & 37.5 & 28.5   & -       & 120.9 & 21.7  \\
VipCap \cite{kim2025vipcap-58}       & Retrieval & -    & 37.7 & 28.6   & -       & \textbf{122.9} & \underline{21.9}  \\ \hline
ClipCap*(base) \cite{mokady2021clipcap-62} & Enc-Dec         & 73.4 & 32.6 & 27.5   & 55.4    & 111.2 & 20.9  \\
RORPCap(Ours)              & Retrieval & 77.5 & 36.7 & \underline{28.8}   & 57.5    & 120.5 & \textbf{22.0}  \\ \hline
\end{tabular}}}
\label{T1}
\end{table}

In Table~\ref{T1}, detector-based models use pre-trained detectors (e.g., Faster R-CNN) to extract visual features, whereas RORPCap bypasses this computationally intensive step. By comparing with Trans. \cite{cornia2020meshed-52} and CSA-Net \cite{li2023csa-13}, we observe that the results of RORPCap are comparable to those of detector-based models in certain evaluation metrics. This indicates that the RORPCap method can effectively extract critical visual information from images without relying on complex detection pipelines. Moreover, although GCN-based models show significant performance gains, the RORPCap method remains competitive without employing GCN to construct image features. Compared with PyAtt\cite{chen2022improving-53} and SSVSGOAR+S\cite{xiao2024sentinel-56}, the RORPCap method still achieves advantages in semantic-level evaluation metrics like CIDEr and SPICE. It suggests that RORPCap can capture the semantic relations in the image and generate semantically coherent captions without relying on GCN. Furthermore, unlike traditional retrieval methods like SMALLCAP\cite{ramos2023smallcap-8} and EXTRA\cite{ramos2023retrieval-9}, which directly utilize retrieved complete sentences, the RORPCap method extracts key information from the retrieved sentences before generating captions. The comparative experimental results show that the key information extraction method is also highly competitive in semantic accuracy and generation quality compared to existing advanced retrieval-based models. The proposed RORPCap not only avoids relying on complex detectors and GCN structures, but also surpasses many existing image captioning methods in several key evaluation metrics, proving its effectiveness as an alternative solution.

\subsubsection{\textbf{Training Time Analysis}}

We also measured the training time and the number of training parameters to further validate the RORPCap performance, as summarized in Table~\ref{T2}. Although large-scale end-to-end detector-based models such as GRIT* \cite{nguyen2022grit-33} and CSA-Net* \cite{li2023csa-13} (both using trainable DETR) perform superior performance metrics, the number of training parameters and the training time of them are extremely large. 
\begin{table}
\caption{Performance comparison with other models on MS-COCO offline Karpathy test split. *: Detector is trainable. Bold denotes the best results, while underline represents the second-best results. The results obtained by the RTX 4090 are all computed by us.}
\renewcommand\arraystretch{1.25}
\centering
\setlength{\tabcolsep}{0.6mm}{
\resizebox{\linewidth}{!}{
\begin{tabular}{l|cccccc}
\hline
Models                  & B@4($\uparrow$)  & M($\uparrow$) & C($\uparrow$) & S($\uparrow$) & Params($\downarrow$) & Time($\downarrow$)  \\ \hline
Up-Down \cite{anderson2018bottom-50}          & 36.2 & 27.0   & 113.5 & 20.3  & 52M          & 960h(M40)      \\
VLP \cite{zhou2020unified-60}           & 36.5 & 28.4   & 117.7 & 21.3  & 115M         & 48h(V100)      \\
Oscar \cite{li2020oscar-61}         & 36.6 & \textbf{30.4}   & 124.1 & 23.2  & 135M         & 74h(V100)      \\
Trans.\cite{cornia2020meshed-52}        & 35.8 & 28.0   & 115.4 & 21.1  & 38M          & 5.2h(RTX4090)  \\
Trans.* \cite{cornia2020meshed-52}       & \underline{39.5} & 29.2   & 127.9 & \underline{23.3}  & 160M         & 45h(RTX4090)   \\
COS-Net \cite{li2022comprehending-26}       & 38.1 & 29.2   & 124.0 & 22.3  & 78M          & 10.1h(RTX4090) \\
GRIT \cite{nguyen2022grit-33}       & 36.5 & 28.2   & 117.6 & 21.2  & 39M          & 18h(RTX4090) \\
GRIT* \cite{nguyen2022grit-33}       & 40.0 & 29.6   & \underline{128.3} & \underline{23.3}  & 160M          & 47h(RTX4090) \\
CSA-Net \cite{li2023csa-13}       & 38.3 & 28.5   & 120.0 & 21.8  & 53M          & 8h(RTX4090)    \\
CSA-Net* \cite{li2023csa-13}      & \textbf{40.3} & \underline{29.5}   & \textbf{130.2} & \textbf{23.6}  & 174M         & 48h(RTX4090)   \\
SMALLCAP \cite{ramos2023smallcap-8}       & 36.2 & 27.6   & 117.7 & 21.0  & \textbf{7M}           & \underline{4.4h}(RTX4090)  \\
LSTNet \cite{ma2023towards-51}        & 36.6 & 28.3   & 116.1 & 21.2  & \underline{38M }         & 5.5h(RTX4090)  \\
ClipCap(base) \cite{mokady2021clipcap-62} & 32.6 & 27.5   & 111.2 & 20.9  & 43M         & 6h(RTX4090)    \\
RORPCap(Ours)                & 36.7 & 28.8   & 120.5 & 22.0  & 116M         & \textbf{2.6h}(RTX4090)  \\ \hline
\end{tabular}}}
\label{T2}
\end{table}

As shown in Table~\ref{T2}, a significant advantage of the RORPCap is the shortest training time for comparable performance metrics. The earlier lightweight model ClipCap\cite{mokady2021clipcap-62} and the recently proposed advanced lightweight model SMALLCAP\cite{ramos2023smallcap-8} both use GPT-2 as their language model. RORPCap has a larger number of parameters due to fine-tuning the language model. Notable, RORPCap reduces training time by nearly two hours and outperforms SMALLCAP. RORPCap also outperforms image captioning models with frozen detectors, while requiring less training time. Although our metrics slightly fall short compared to detector-based end-to-end models, the RORPCap model offers significant advantages in parameter count and training time.

\subsubsection{\textbf{Results on nocaps dataset}}

To further evaluate the generalization of RORPCap, the Table~\ref{T3} shows test results on the nocaps dataset, where we only utilize information retrieved from the MS-COCO dataset for training and testing on the nocaps dataset. The nocaps dataset is used to measure generalization to unseen classes and concepts. As can be seen, whether compared to detector-based models or retrieval-based models, RORPCap achieves competitive performance across the three different parts of the nocaps dataset. The RORPCap model can extract objects beyond the 80 categories in the MS-COCO dataset, which allows RORPCap to have a strong and broad conceptual labeling capability. This is crucial for images containing novel concepts, as it is helpful to generate open-form descriptions.
\begin{table}
\caption{Zero-shot performance on the nocaps validation split. Bold denotes the best results, while underline represents the second-best results, and (-) indicates unevaluated scores. *: Results computed by us.}
\renewcommand\arraystretch{1.25}
\centering
\setlength{\tabcolsep}{0.5mm}{
\resizebox{\linewidth}{!}{
\begin{tabular}{l|cccccccc}
\hline

\multirow{2}{*}{Models} & \multicolumn{2}{c}{In-domain} & \multicolumn{2}{c}{Near-domain} & \multicolumn{2}{c}{Out-of-domain} & \multicolumn{2}{c}{Overall} \\ \cmidrule(r){2-3}\cmidrule(r){4-5} \cmidrule(r){6-7} \cmidrule(r){8-9} 
                       & C         & S         & C          & S          & C           & S           & C        & S        \\ \hline
Up-Down \cite{anderson2018bottom-50}        & 78.8          & 11.6          & 57.7           & 10.3           & 31.3            & 8.3             & 55.3         & 10.1         \\
Oscar \cite{li2020oscar-61}          & 79.6          & \underline{12.3}          & 66.1           & \underline{11.5}           & 45.3            & \textbf{9.7}             & 63.8         & \underline{11.2}         \\
Trans. \cite{cornia2020meshed-52}         & 78.0          & 11.0          &-                &-                & 29.7            & 7.8             & 54.7         & 9.8          \\
M2 \cite{cornia2020meshed-52}             & \textbf{85.7}          & 12.1          &-                &-                & 38.9            & 8.9             & 64.5         & 11.1         \\
I-Tuning \cite{luo2023tuning-63}       & 83.9          & \textbf{12.4}          & 70.3           & \textbf{11.7}           & 48.1            & \underline{9.5}             & 67.8         & \textbf{11.4}         \\
SMALLCAP* \cite{ramos2023smallcap-8}       & 80.2          & 11.8          & \textbf{77.0}           & 11.4           & \textbf{64.3}            & 9.2             & \textbf{75.2}         & 10.7         \\
EntroCap \cite{yan2025entrocap}       & 62.5          & 10.4          & 64.5           & 10.0           & 67.5            & 9.0             & 67.0         & 9.5         \\
ClipCap*(base) \cite{mokady2021clipcap-62}       & 76.7          & 12.0          & 65.5           & 10.9           & 48.1            & 9.2             & 64.4         & 10.8         \\
RORPCap(Ours)                & \underline{84.4}          & \textbf{12.4}          & \underline{73.8}           & \textbf{11.7}           & \underline{55.2}            & 9.4             & \underline{70.3}         & 11.1     \\ \hline   
\end{tabular}}}
\label{T3}
\end{table}

\subsection{\textbf{Futher Analysis}}
\begin{figure*}[h]
\centerline{\includegraphics[width=7.5cm]{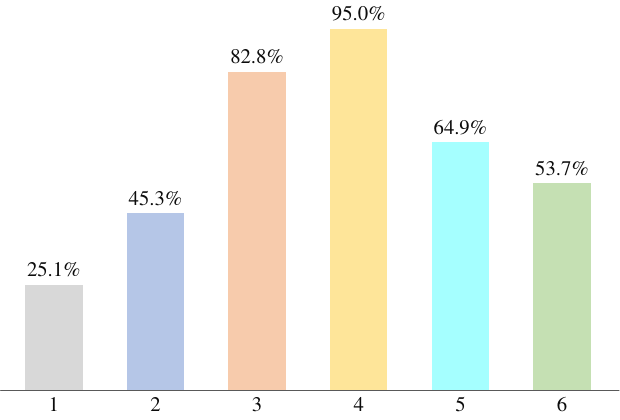}}
\caption{Different proportion of object and relation words extracted after retrieval compared to the words used in the GT.} 
\label{fig5}
\end{figure*}

The proportion of object and relation words extracted by OREM compared to the words used in the GT is analyzed. The GTs are manually annotated, and analyzing the proportion of words can help us determine whether the OREM effectively extracts key words. We calculated the distribution of objects in the MS-COCO dataset, where sentences containing six or fewer objects account for 99.7\%. In other words, at most six objects are sufficient to describe the input image. Through these statistical results, we can evaluate the performance of the retrieval module and explore its feasibility as an image description model based on the detector and GCN, as shown in Fig.~\ref{fig5}.

 As illustrated in Fig.~\ref{fig5}, Column 1 and Column 2 respectively represent the retrieved object and relation words that appear in the GT sentences. Column 3 and Column 4 respectively indicate that at least three and one retrieved objects and relation words appear in the GT sentences. We believe that in the five description sentences matching each image, objects that appear at least three times and relation words that appear at least twice are the primary objects and relations that people focus on when describing the image. Column 5 counts the cases where at least two retrieved objects appear at least three times in the five description sentences. Column 6 counts the cases where at least one retrieved relation word appears at least three in the five description sentences. These data indicate that our retrieval model can identify the key object and main relation words when describing an image, thereby reducing redundant information, lowering computational costs, and accelerating model training.

\subsection{\textbf{Ablation Study}}

In this section, we conduct an ablation study to investigate how each component of RORPCap influences the overall performance on the MS-COCO dataset.

\subsubsection{\textbf{The impact of each component}}
\begin{table}
\caption{Ablation study for RORPCap on MS-COCO Karpathy test split. Prompt: The object and relation words derived from OREM are filled into the template to form the prompt. Mamba: Mapping network built based on Mamba. Transformer: The mapping network based on Transformer constructed in ClipCap \cite{mokady2021clipcap-62}. GPT-2 tuning: Fine-tuning GPT-2, all of them use GPT-2-small as a language model.}
\renewcommand\arraystretch{1.25}
\centering
\setlength{\tabcolsep}{6pt}
\begin{tabular}{c|cccc|ccc}
\hline
\multicolumn{1}{c|}{\#} & Prompt & Mamba & Transformer & GPT-2 tuning & B@4  & C     & S    \\ \hline
1                       &        & \checkmark     &             &              & 31.1 & 103.9 & 19.9 \\
2                       &        &       & \checkmark           &              & 32.1 & 107.5 & 20.3 \\
3                       & \checkmark      &       & \checkmark           &              & 31.6 & 112.9 & 21.7 \\
4                       &        & \checkmark     &             & \checkmark            & 32.5 & 110.1 & 20.3 \\
5                       & \checkmark      & \checkmark     &             &              & 28.2 & 101.4 & 21.6 \\
6                       & \checkmark      &       & \checkmark           &              & 32.5 & 111.6 & 21.5 \\
7                       & \checkmark      &       & \checkmark           & \checkmark            & 33.7 & 116.5 & 21.9 \\
8                       & \checkmark      & \checkmark     &             & \checkmark            & 36.7 & 120.5 & 22.0 \\ \hline
\end{tabular}
\label{T4}
\end{table}
In Table~\ref{T4}, we conduct ablation experiments on the key components of the model. \#1 indicates that using Mamba as the mapping network, along with the language model GPT-2 can still output meaningful sentences. \#1 and \#2 validate the performance differences between the Transformer model and the Mamba model in constructing the mapping from the image domain to the text domain. \#3 represents the case where we add text prompt information to the baseline model ClipCap\cite{mokady2021clipcap-62}, the experimental results clearly show that the prompt information significantly improves the experimental performance. This indicates that the OREM can be transferred to other models and effectively enhance the description performance, which is one of the contributions to the image captioning field. ClipCap concludes that fine-tuning the language model is unnecessary when using the Transformer architecture. The ablation experiments \#4 and \#8 demonstrate that fine-tuning the language model is necessary when using the Mamba architecture, and the results are better than those of the mapping network using the Transformer architecture. When using prompt information, fine-tuning the language model is necessary for both architectures. The experiments further conclude that the Mamba architecture can lead to better results when combined with prompts and language model fine-tuning.

\subsubsection{\textbf{The impact of two types of words}}

\begin{table}
\caption{Impact of Two Types of Words.}
\renewcommand\arraystretch{1.25}
\centering
\setlength{\tabcolsep}{12pt}
\begin{tabular}{c|cc|ccc}
\hline
\# & Objects & Relations & B@4  & C & S \\ \hline
1  &         &           & 32.5 & 110.1 & 20.3  \\
2  & \checkmark       &           & 35.4 & 117.2 & 21.6  \\
3  &         & \checkmark         & 33.9 & 111.2 & 20.4  \\
4  & \checkmark       & \checkmark         & 36.7 & 120.5 & 22.0  \\ \hline
\end{tabular}
\label{T5}
\end{table}
We also perform an experimental analysis of two types of words to explore their impact on image captioning generation. The key to image captioning generation lies in identifying objects and their relations in the image, which contributes to richer and more consistent descriptions of the image. As shown in Table~\ref{T5}, we find that the object prompt has the maximum gain for caption generation, e.g., 6.9\% CIDEr score, which achieved 117.2\% CIDEr score, still outperforming some detector-based image captioning models, as shown in Table~\ref{T1}. The relation words as prompts resulted in a smaller improvement, e.g., 1.1\% CIDEr score. Finally, when the complete prompt is integrated into the model, its performance improves significantly and reaches 10.4\% CIDEr score. The possible reason is that the accuracy of object and relation semantics obtained solely through mapping networks is inadequate. In other words, using only objects as prompts does not capture the correct relations, resulting in limited improvement. Using only relation words as prompts leads to the model failing to accurately identify objects and resulting in insufficient enhancement.

\subsubsection{\textbf{The impact of different Types of Prompt Template}}

\begin{table}
\caption{Impact of different Types of Prompt Template.}
\renewcommand\arraystretch{1.25}
\centering
\setlength{\tabcolsep}{12pt}
\begin{tabular}{c|ccccc}
\hline
Prompt Template & B@4  & M & R & C & S \\ \hline
\#1               & 35.6 & 28.1   & 57.0    & 117.3 & 21.6  \\
\#2               & 36.4 & 28.6   & 57.3    & 119.4 & 21.8  \\
\#3               & 36.7 & 28.8   & 57.5    & 120.5 & 22.0  \\ \hline
\end{tabular}
\label{T6}
\end{table}
Besides the template proposed in the paper, we also explore other templates for prompting, including: \#1:  "\textit{${object_1}$}, \dots, \textit{${object_n}$}. \textit{${relation_1}$}, \dots, \textit{${relation_j}$}. \{mapping network output and GT\}". \#2 "\textit{a photo of} \textit{${object_1}$}, \dots, \textit{${object_n}$}. A\textit{ photo contains the relations of} \textit{${relation_1}$}, \dots, \textit{${relation_j}$}. \textit{Its caption is} \{mapping network output and GT\}". \#3: "\textit{a photo contains objects:} \textit{${object_1}$}, \dots, \textit{${object_n}$}, \textit{and the relations are} \textit{${relation_1}$}, \dots, \textit{${relation_j}$}. \textit{Its caption is} \{mapping network output and GT\}". We conduct experiments on the MS-COCO dataset, as shown in Table~\ref{T6}. Experimental results confirm that \#3 performs better, and we choose it as our final prompt template.

\subsubsection{\textbf{Parameter Analysis}}

\renewcommand{\figurename}{Fig.}
\begin{figure}[h]
\centering
\centerline{\includegraphics[width=9cm]{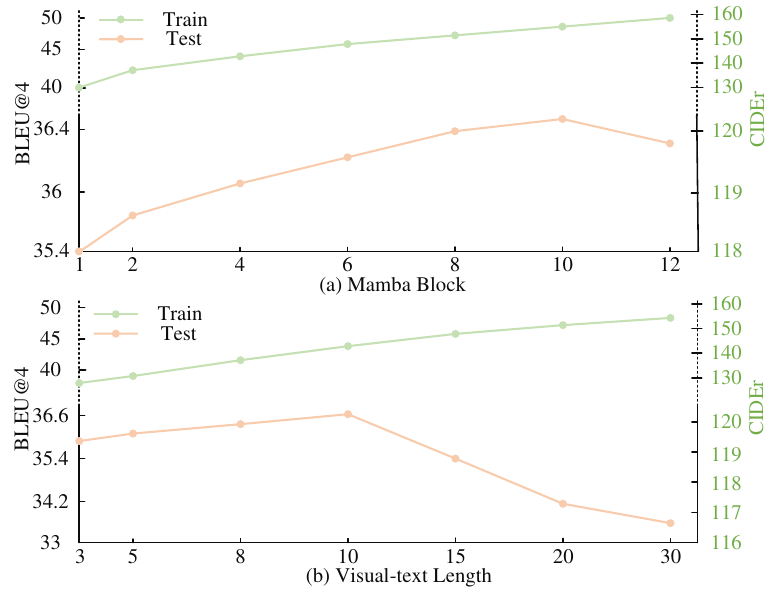}}
\caption{Effect of the Mamba Block and Visual-text length on the captioning performance over the MS-COCO dataset. For each Mamba Block and Visual-text length, we report the BLEU@4 (black) and CIDERr (green) scores over the test (orange line) and train (green line) sets.}
\label{fig6}
\end{figure}
We analyze two main parameters of the model, i.e., Mamba block and visual-text length. As shown in Fig.~\ref{fig6}, we can see that increasing the number of Mamba blocks and Visual-text length improves model performance up to a certain point, after which the model's performance starts to decline. With the increase in the number of Mamba blocks and Visual-text length, the number of trainable parameters in the model also increases, leading to overfitting on the training set and reduced performance on the test set. On the other hand, if the number of Mamba blocks and the Visual-text length has not reached a certain value, the results are relatively poor, as the model is not expressive enough.

\section{Qualitative Results} \label{sec:5}
\begin{figure}[h]
\centering
\includegraphics[width=1.0\linewidth]{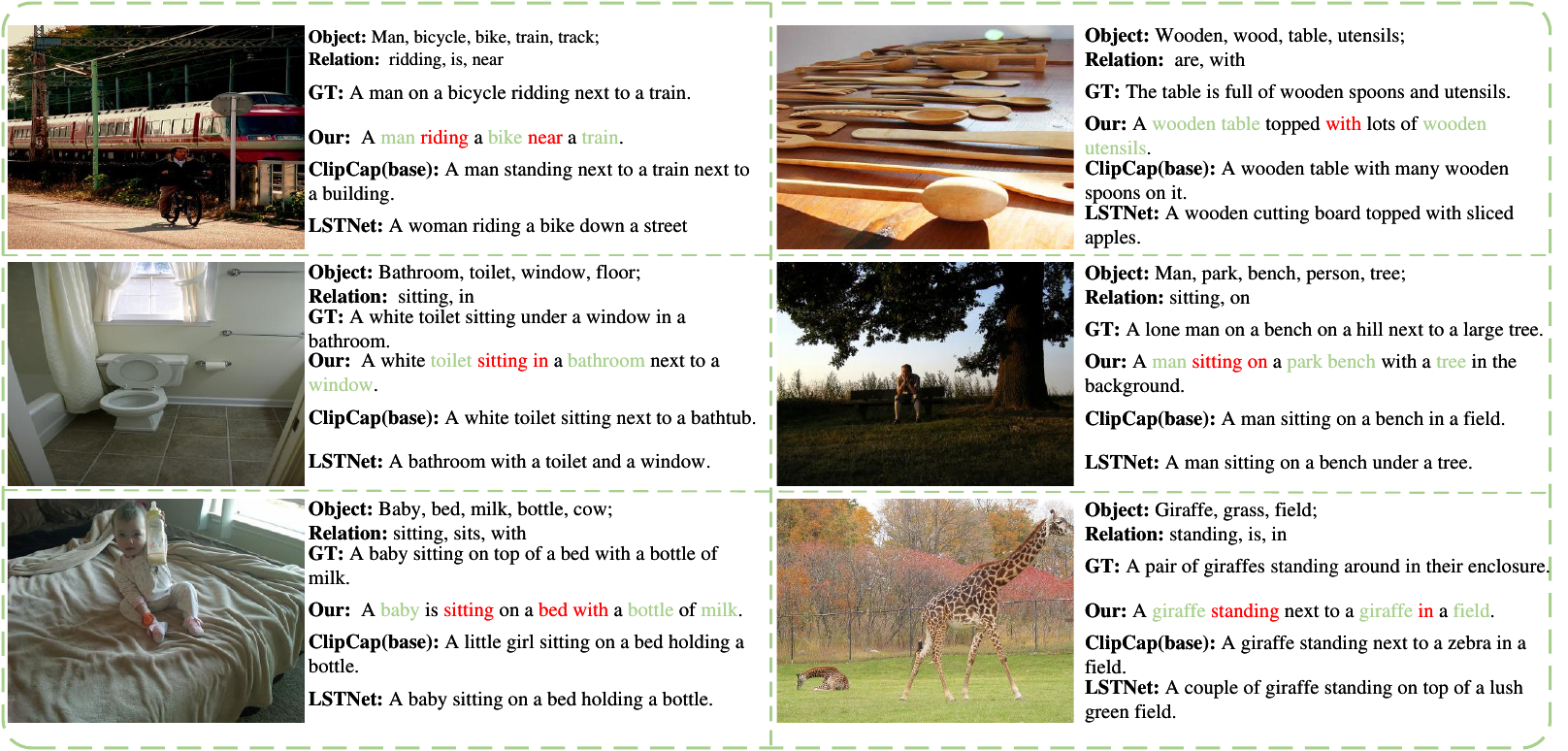}
\caption{Examples of captions generated by different methods. "Object" and "Relation" represent the object and relation words extracted from the retrieval sentences using OREM. As well as generated captions by the RORPCap model, the base model, and the recent detector-based image captioning model LSTNet. The green and red words represent the generated object words and relation words, which match the words using OREM.}
\label{fig7}
\end{figure}

We showcase several qualitative results of RORPCap to qualitatively demonstrate its effectiveness. In Fig.~\ref{fig7}, the results of RORPCap, the baseline model ClipCap, the detector-based model LSTNet, and human-annotated GT sentences are compared. We highlight the correct object and relation words in our model with green and red, respectively. Overall, it is clear that all three methods are capable of identifying objects and their relations in the image to generate linguistically coherent descriptions. However, upon examining the semantic-related objects and relations, RORPCap can effectively capture more relevant semantic words through OREM. It also outperform both the baseline model and the detector-based model LSTNet. These object and relation words are crucial for conveying meaningful image descriptions. For example, it is evident that RORPCap identifies more objects (e.g., man, utensils, bike) and more accurate relations (e.g., riding, sitting, in, with). It further demonstrates that RORPCap can provide richer object and relation words as cues for image descriptions, even without using detectors or GCN.
\renewcommand{\figurename}{Fig.}

\section{Conclusion} \label{sec:6}
In this paper, a retrieval-based image captioning approach RORPCap is proposed to extract object and relation words in images. Traditional image captioning models suffer from such as redundant detection information, difficulties in constructing GCN, and high training costs. The designed OREM in the RORPCap extracts object and relation words from the retrieved sentences and builds a prompt template for input, enabling the model to effectively extract the main descriptive objects and relations in images without the need for detectors or GCNs. The RORPCap model achieves comparable descriptive accuracy to traditional image description models based on detector and GCN, while significantly reducing training time and costs. It is the model with the shortest training time for comparable performance metrics. Future research could explore how to leverage retrieval-based methods to address more challenging visual-to-text tasks such as visual question answering, video description, and scene graph generation. This could help further reduce training costs and model complexity.

\bibliographystyle{apalike}
\bibliography{references}  


\end{document}